\documentclass[a4paper]{article}

\usepackage{INTERSPEECH2021}
\usepackage{stfloats}

\title{Team MTS @ AutoMin 2021: An Overview of Existing Summarization Approaches and Comparison to Unsupervised Summarization Techniques}
\name{Olga Iakovenko$^1$, Anna Andreeva$^2$, Anna Lapidus$^1$, Liana Mikaelyan$^1$}
\address{
  $^1$MTS AI\\
  $^2$ActiveBC}
\email{osyakov4@mts.ru, rogotulka@gmail.com, aalapidu@mts.ru, lvmikaye@mts.ru}

\begin{document}

\maketitle
\begin{abstract}
  Remote communication through video or audio conferences has become more popular than ever because of the worldwide pandemic. These events, therefore, have provoked the development of systems for automatic minuting of spoken language leading to AutoMin 2021 challenge. The following paper illustrates the results of the research that team MTS has carried out while participating in the Automatic Minutes challenge. In particular, in this paper we analyze existing approaches to text and speech summarization, propose an unsupervised summarization technique based on clustering and provide a pipeline that includes an adapted automatic speech recognition block able to run on real-life recordings. The proposed unsupervised technique outperforms pre-trained summarization models on the automatic minuting task with Rouge 1, Rouge 2 and Rouge L values of 0.21, 0.02 and 0.2 on the dev set, with Rouge 1, Rouge 2, Rouge L, Adequacy, Grammatical correctness and Fluency values of 0.180, 0.035, 0.098, 1.857, 2.304, 1.911 on the test set accordingly
\end{abstract}

\section{Introduction}

In the modern society, remote meetings have become an essential component of various team activities, increasing the demand for extraction of the most important points of a meeting. Besides that, minuting of meetings, both online and offline, has always been a useful practice. Meeting minutes can help teams consolidate certain ideas and plans, without dissolving into oblivion, while also serving as a proof that the team is spending their time resources effectively.
Previous to the Automatic Minutes Challenge, there has been a limited number of similar activities, one of which was a session on speech summarization took place in 2006  \cite{Richmond2006INTERSPEECH2}. Since then, there has been little research on speech summarization, summing up to 153 papers \cite{Rezazadegan2020AutomaticSS}, even though the Natural Language Processing (NLP) methods have undergone a considerable improvement. Advances in NLP techniques have contributed to more robust and accurate text summarization methods \cite{Raffel2020ExploringTL,Rothe2020LeveragingPC,Dong2019UnifiedLM}.

The growing necessity for online communication as the development of new approaches towards text summarization resulted in the re-emerging of interest for automatic minuting and, consequently, the Automatic Minutes challenge was launched \cite{Ghosal2021}.
This paper illustrates the results of the research that team MTS has carried out in the process of making a solution for the Automatic Minutes challenge. The following paper consists of six major parts:

\begin{itemize}
	\item Introduction, covering the challenge motivations;
	\item Background review discussing main approaches used in speech summarization;
	\item Description of the datasets used and of the designed approaches;
	\item Experiment procedure and results;
	\item Conclusion and further work.
\end{itemize}
In this project we aim to achieve three main goals. First, we wish to analyze the existing pre-trained summarization models and compare their performance in summarization of manually transcribed audio recordings. Second, we propose a custom unsupervised approach for summarization of English texts. Last but not least, we adapt our summarization module for multi-channel meeting audio recordings and made the pipeline open-source.

\section{Background}

\begin{figure}[h]
	\centering
	\includegraphics[scale=0.35]{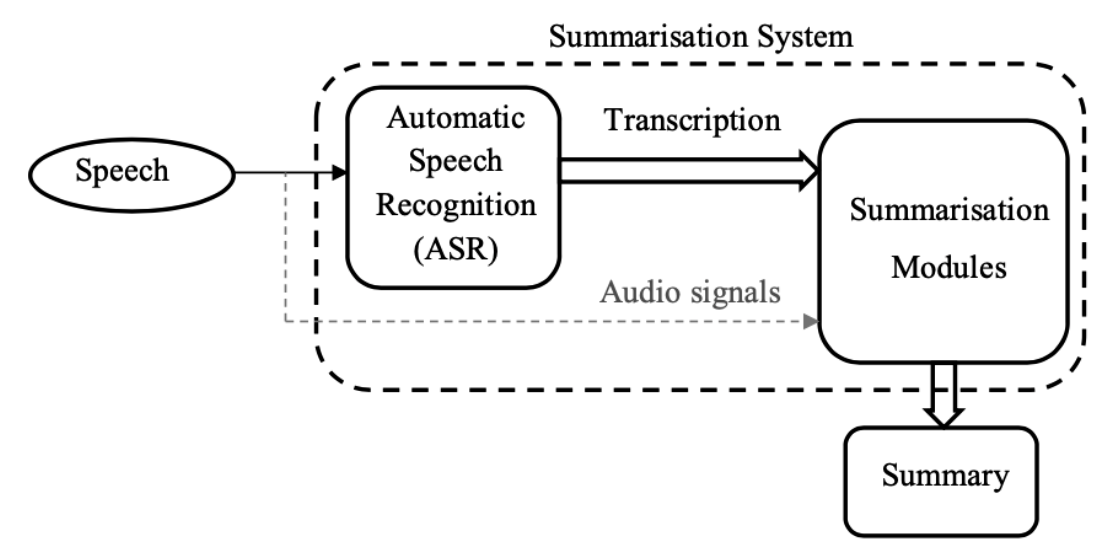}
	\caption{Conventional structure of a speech summarization system}
	\label{fig:speech}
\end{figure}

Traditionally speech summarization tasks or automatic minuting of speech are tackled using a component-based approach. A typical speech summarization system consists of two components: an ASR component and a summarization component (Figure~\ref{fig:speech}). Some additional acoustic features can be separately extracted from audio and propagated directly to the summarization module.

There are two main approaches for speech and text summarization: extractive and abstractive. Extractive summarization identifies informative utterances or sentences that best describe the main focus of a meeting or text. Such summarizations may be post-processed to form a more coherent and precise text. Abstractive summarization, on the other hand, uses generative machine learning algorithms to generate a summary based on input speech or text. An abstractive summary may be a paraphrase of the input data containing the main idea of the text.

For the past 15 years there has been a moderate amount of research on the topic of automatic meeting speech summarization and note-taking. Extractive methods make use of algorithms for utterance ranking by “notability” or “informativeness”, similar to the term “meaningfulness” that we have presented in the following paper. For example, first works on this topic include approaches based on top-n informative words or phrases \cite{Murray2007TermWeightingFS,Szaszak2016}, Naïve Bayes classifier and Decision Trees \cite{Banerjee2009}, Logistic Regression \cite{Murray2008MetaCF, Basu2008ScalableSO}, SVM \cite{Tokunaga2014MultipartyCS}, skip-chain conditional random field \cite{Galley2006}. Some works investigate the correlation between text summarization and speech summarization \cite{Murray2008SumConv}, while others look into the effect of automatic sentence segmentation on speech summarization \cite{Liu2008ImpactOA}. Apart from speech summarization, research on multimodal summarization was carried out by Metze et al \cite{Metze2012BeyondAA} where both video and audio data were utilized for the task.
The aforementioned speech summarization makes use of different features such as:

\begin{itemize}
	\item Lexical – grammatical, syntactical and semantical information about words and their relationships.
	\item Acoustic – tone and emotion of the utterance.
	\item Structural – position of words and utterances respect to t.
	\item Relevance – relevance scores, obtained through preliminary text vectorization and consecutive similarity computations of the whole text and utterances/sentences
\end{itemize}

\section{Experiments}

\subsection{Datasets}

The main dataset we employ in our experiments is Automin-confidential-data. 

\subsubsection{Automin-confidential-data}

We employ a custom private dataset composed for the Automatic Minutes challenge to train and fine-tune summarization models. The dataset contains textual data of manual transcriptions for meeting recordings. 

The dataset includes 6 subsets in total, containing two subsets in English and Czech language for each of the tasks:

\begin{itemize}
	\item Task A: transcriptions and corresponding minutes (can be several minutes for one meeting).
	\item Task B: transcriptions and minutes, where a pair of a transcription and a minute can be descriptions of the same meeting or two different meetings.
	\item Task C: two minutes, which can describe two different meetings, or the same meeting.
\end{itemize}

Names of web pages, companies and people in both transcription and minutes are anonymized and replaced by such placeholders as “organization1” or “person1”. Inconsistent capitalization and punctuation are present in both transcription and minutes. Some minutes contain a header, which is ignored during evaluation. Some of the phrases in transcriptions provide the speaker of a phrase. Open-source examples of the dataset are available via link \cite{AutoMinData}.

We follow the structure of Automin dataset to use the same train/dev/test split, combining two parts of test set (Table~\ref{tab:data}).

\begin{table}[th]
  \caption{Statistics for train, test, dev sets of Automin dataset}
  \label{tab:data}
  \centering
  \begin{tabular}{l r  r   r  }
    \toprule
     Train  &  Test  &  Dev  \\
     \midrule
     85  &  28  &  10  \\

    \bottomrule
  \end{tabular}
  
\end{table}

\subsubsection{ICSI}

In addition to the automin-confidential-data, we use ICSI Meeting corpus \cite{Janin2003TheIM} to evaluate summarization pipeline on the audio data.  ISCI dataset contains transcripts for audio meeting recordings. Specifically, we utilize the individual channel streams to compute the result of summarization for the meetings data.

\subsubsection{AMI}

We also use AMI Meeting Corpus \cite{McCowan2005TheAM} textual data for the adaptation of the speech summarization pipeline. We employ the AMI manual transcriptions of phrases pronounced in the recorded meetings to train Language Model (LM) for ASR component.

\subsection{Algorithms}

\begin{figure}[h]
	\centering
	\includegraphics[scale=0.45]{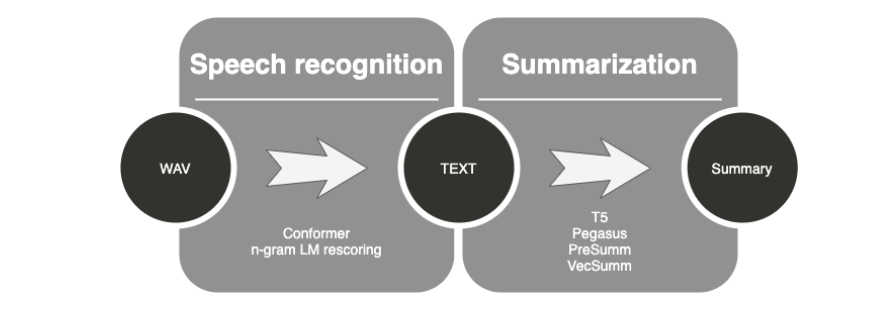}
	\caption{Developed speech summarization pipeline}
	\label{fig:pipeline}
\end{figure}

Our pipeline consists of two components: speech recognition and summarization (Figure~\ref{fig:pipeline}).

Transfer learning that involves training a model on a large unlabeled dataset and fine-tuning it to downstream tasks has become a major component in many text-related tasks. Therefore, in this project, we focus our attention on three pre-trained algorithms for text summarization and compared the pre-trained algorithms with a custom unsupervised approach. We consider pre-trained models that can be potentially fine-tuned for the task of speech summarization and are designed to generate abstractive summaries. As for the custom unsupervised approach that we name VecSumm, we chose USE vectorization for effective summarization of syntagms and Affinity Propagation clustering for reverse cluster center selection. From here onwards we describe all of the algorithms that are applied to the English version of the Task A dataset.

\subsection{Speech Recognition}

The speech recognition component of our system accepts a collection of 16 kHz audio files, each of which represent a separate channel (speaker) in the meeting. Audio batch is then processed by a Voice Activity Detector (VAD) to remove silent segments. Here we employ Silero VAD \cite{Silero}. The results of VAD processing are then being ordered in correspondence to their appearance in the meeting to reconstruct the flow of the discussion.

Each speech segment is then fed into a speech recognition module. Here we adopt a pre-trained Conformer model, provided by NeMo \cite{NeMo}. Output of the Conformer is rescored by an n-gram LM through beamforming. We constructed a custom n-gram LM for rescoring using the manual transcriptions of AMI corpus. The output of the speech recognition module is then passed through to the summarization module without any punctuation restoration.

If our system encounters textual data instead of an audio, initial transcriptions are preprocessed to resemble an output of the speech recognition module: input text is split into sentences, punctuation is removed and all characters are transformed to lowercase.

\subsection{Text Summarization}

In our experiments we explore multiple deep learning-based text summarization techniques and then propose an alternative custom unsupervised technique. 

\subsubsection{Text-To-Text Transfer Transformer}

Text-To-Text Transfer Transformer (T5), trained on Colossal Clean Crawled Corpus (C4), has shown a state-of-the-art performance on many NLP benchmarks \cite{Raffel2020ExploringTL}. We use the T5 model to produce summaries for small passages, extracted from the transcriptions. Moreover, we attempt to fine-tune the T5 model for the exact task of automatic minuting.

We start by adapting Automin-confidential-data for the T5 fine-tuning procedure. Because we do not know which part of the transcription corresponds to given summaries, the first step is to implement an automatic way to perform such correspondence. That is, for each transcription and summaries file, we need to obtain $ (x_{i}, y_{i}) $ pairs, where $ x_{i} \in X $ is a passage from a transcription file $ X $ and $ y_{i} \in Y $ is a summary from a summary file $ Y $, while $ i $ and $ j $  are indices of each piece of text in the transcription and summaries file respectively. 

To accomplish this, we break both the transcription and summaries files into passages and feed combinations into the T5 model recording the obtained scores for each pair. We then choose pairs with lower losses since they are more likely to be valid pairs.
During training we load the data in this manner, at each epoch reloading the data with the current model parameters so that the loaded transcription and summary pairs become more valid. Thus, as the performance of the model improves, the loaded pairs quality also improves, in turn helping to improve the model as well.

Example of generated summary (Table~\ref{tab:T5_example}):

\begin{table}[th]
  \caption{T5 generated text}
  \label{tab:T5_example}
  \centering
  \begin{tabular}{|p{7cm}|}
    \toprule
     so from this monday we can actually go out even if it's not like the necessity
     it the meetings of. up to ten people are allowed. ha. so here in location1 we think that we have to wait until june for the free circulation of people. starting from the fourth of june we are allowed to reach our family. \\
    \bottomrule
  \end{tabular}
\end{table}

\subsubsection{Pegasus}

Among other pretrained models, Pegasus model \cite{Zhang2020PEGASUSPW} demonstrated state-of-the-art performance on abstractive summarization while fine-tuned on small specific datasets. Pegasus is a Transformer-based model pretrained on C4 dataset of web-pages and HugeNews dataset of news articles. During training for abstractive summarization task the most informative sentences were removed from input text, and the model was trained to generate a summary close to the hidden sentences. 

Pegasus-large model is used as a pretrained model for this task. The model contains 16 encoder and 16 decoder layers, 16 encoder and 16 decoder attention heads, maximum input length for the model is 1024.

Example of generated summary (Table~\ref{tab:Pegasus_example}):

\begin{table}[th]
  \caption{Pegasus generated text}
  \label{tab:Pegasus_example}
  \centering
  \begin{tabular}{|p{7cm}|}
    \toprule
     so from this monday we can actually go out even if it's not like the necessity. but we think that we have to wait until june for the free circulation of people. and fortunately starting from the fourth of june we are allowed to reach our family. well the rules changed here and since this monday we can't go out even if it is not like the most necessary groceries or stuff like this and. yes but if you are in the city then you have to somehow get to the forest so. probably yes because organization8person8 and person14 said that they have the call so they can't join ours. one in may and one in june and. \\
    \bottomrule
  \end{tabular}
\end{table}

\subsubsection{PreSumm}

PreSumm \cite{Liu2019TextSW} is an implementation of Bert model fine-tuned for summarization task. It was trained on large corpuses of public dataset (CNN/DailyMail, the New York Times Annotated Corpus and XSum).

Before being fed to the pre-trained PreSumm algorithm, the conversation transcriptions are preprocessed, namely interjections and frequent words are removed. Afterwards, to be able to operate with long sequences (e.g., the transcriptions from the target dataset automin-confidential-data) we split each example from the dataset into chunks of no more than 512 words. We do not split up a sentence or phrase into two different parts. Then we run all the chunks through the pre-trained PreSumm model and get the summarized parts. The process is being repeated until we get the summarization text length no more than 10\% of the length of the original file.

Example of generated summary (Table~\ref{tab:Presumm_example}):

\begin{table}[th]
  \caption{Presumm generated text}
  \label{tab:Presumm_example}
  \centering
  \begin{tabular}{|p{7cm}|}
    \toprule
     people can wait until june for the free circulation of people ( person13 ) and my family is in bolzano which is pretty near around 50km ( person5 ) so you ca n't go to the forest or something ? ( person9 ) is not coming ( person6 ) , so we have every partner- a partner from everywhere this is again one of our regular calls there will be at least one more , in may and one in june , and in may person6 ( person1 ) is the same type of telephone-based list it 's also possible to list your name or not ? you 'll let us know person6 ( person5 ) is the first one so i will put here [ person5 ] plus [ person8 ] $\langle...\rangle$\\
    \bottomrule
  \end{tabular}
\end{table}

\subsubsection{VecSumm}

\begin{figure}[h]
	\centering
	\includegraphics[scale=0.45]{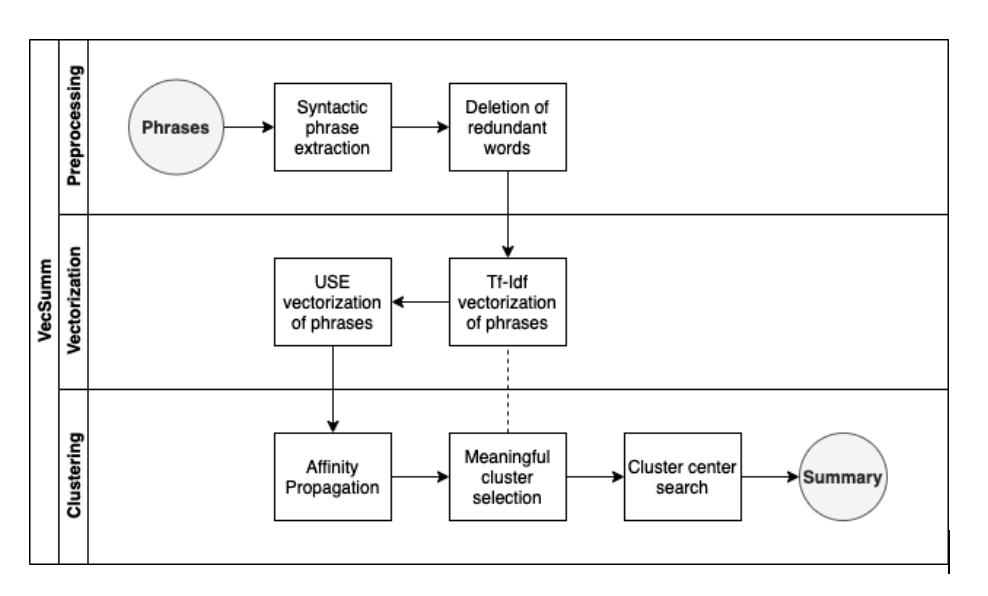}
	\caption{Flow of the VecSumm approach}
	\label{fig:VecSumm}
\end{figure}

As an alternative for the pre-trained methods mentioned above a method based on clustering and vectorization is produced Figure~\ref{fig:VecSumm}.

Let us introduce the following steps as illustrated in Figure~\ref{fig:VecSumm}:

\begin{itemize}
	\item Preprocessing: 
		\begin{itemize}
			\item Syntactic phrase extraction: phrase is split into syntactic phrases by punctuation, while also by conjunctions and subjects of the phrase.
			\item Deletion of redundant words: words are filtered by their part of speech affiliation and by their syntactic role. Adverbial modifiers are removed from the word sequences, as for the parts of speech determinants: interjections and subordinating conjunctions are discarded
		\end{itemize}
		
	\item Vectorization:
		\begin{itemize}
			\item  Tf-Idf scores are computed for words in each syntactic phrase extracted within one meeting. The Tf-Idf scores for words in each phrase are averaged deriving a custom metric sentence Tf-Idf  which refers to the meaningfulness of the syntactic phrase. We computed Tf-Idf scores for n-grams ranging from 1 to 3. %%%% math equation here , where  is the length of the word sequence in a syntactic phrase
			\begin{equation}
  				STFIDF_{d,D} = \frac{1}{N} \sum\limits_{i=0}^{N} tfidf (t_{i}, d, D)
  				\label{eq1}
			\end{equation}
			where $N$ is the length of the word sequence in a syntactic phrase
			\item Universal Sentence Encoder \cite{Cer2018UniversalSE} embeddings are computed for each syntactic phrase initialized with defauld parameters.
		\end{itemize}
		
	\item Clustering:
		\begin{itemize}
			\item Affinity propagation clustering method is applied to all the syntactic phrases. The advantage of this method amongst other clustering methods is that it selects specific examples from the datasets to be centers of clusters. This way we can choose the centers of clusters to be our representative phrase of a cluster. We calculate the clusters with the damping factor of 0.9, we set a maximum possible number of iterations for conversion at a value 1000 and a early stopping criterion of 50 iterations with no change of values.
			\item The meaningfulness of a cluster is calculated as a mean of all the  belonging to the cluster.
			\item Cluster center search: the 10\% most meaningful clusters are determined, written in the chronological order and represented by cluster centers of those clusters
		\end{itemize}
\end{itemize}

\begin{figure}[h]
	\centering
	\includegraphics[scale=0.35]{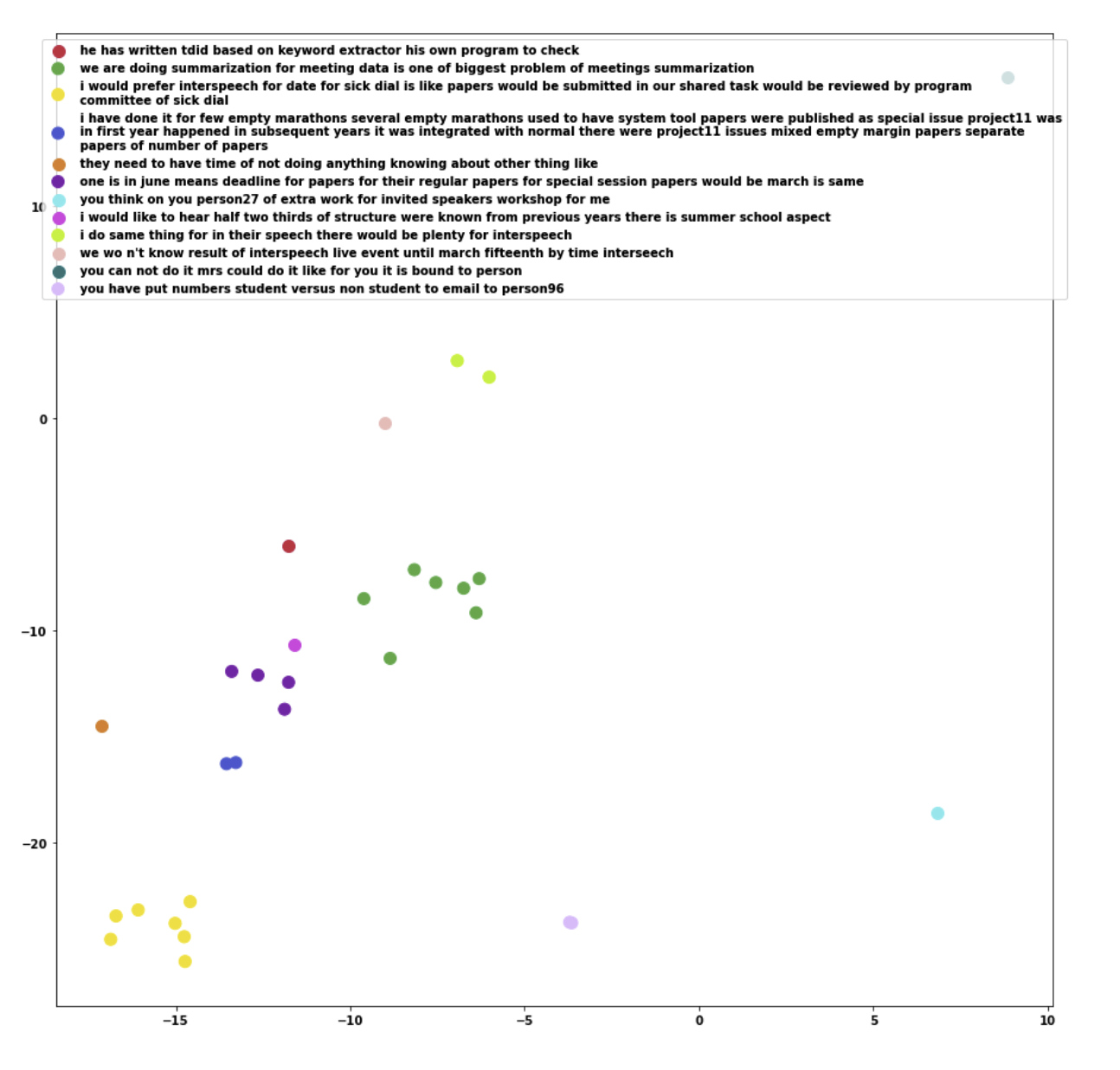}
	\caption{Meaningful clusters and their cluster centers}
	\label{fig:Clusters}
\end{figure}

Figure~\ref{fig:Clusters} illustrates some of the determined clusters from the target data and their centers.

Example of generated summary (Table~\ref{tab:Vecsumm_example}):

\begin{table}[th]
  \caption{Vecsumm generated text}
  \label{tab:Vecsumm_example}
  \centering
  \begin{tabular}{|p{7cm}|}
    \toprule
     we have to wait until june for free circulation of people starting from fourth of june we are allowed to reach our family
people should have little incentive to read it is partner planning to do start anyone
things slept eighth of june sounds fine should be end of review
main responsible for deliverable is organization6 will not be confused from layout of test set
we do include latency sltf does include delay latency wasted effort there is two measures of wasted effort
i have strong preference not to submit my model to ehmm to organizers to run it for one unpublished code
i wanted to mention is forced alignment finds words in sound is not reliable for us it is shifted $\langle...\rangle$ \\
    \bottomrule
  \end{tabular}
\end{table}

\section{Evaluation}

There are two evaluation stages of our system: evaluation during the systems development and final evaluation for the Automatic Minutes challenge. We refer to these evaluations as validation and testing accordingly.

\subsection{Validation}

Each generated minute is compared to the manual minutes of a meeting. All punctuation and capitalization are removed from both generated and manual minutes. For the validation phase we calculate 24 metrics in total:

\begin{itemize}
	\item Rouge (types 1, 2, 4, L, w-1.2, s4, su4)
	\item BERT Similarity
\end{itemize}

Rouge metric measures N-gram overlap between a reference summary and a generated one. Thus, it focuses on matching particular words, rather than word meanings. BERT Similarity, on the other hand, uses BERT's ability to embed words and sentences into vectors representing their meanings, and calculates similarity between those embeddings, providing a more accurate metric for the task where meanings of words are vital.

\begin{table}[th]
  \caption{Validation results}
  \label{tab:validation}
  \centering
  \begin{tabular}{l r  r   r  }
    \toprule
     Method &  Rouge 1 &  Rouge 2 &  Rouge L \\
     \midrule
    $T5$ & $0.14$ & $0.01$ & $0.13$ \\
    $Pegasus$ & $0.2$ & $ \bf 0.03$ & $0.17$ \\
    $VecSumm$ & $ \bf 0.21$ & $0.02$ & $ \bf 0.2$ \\
    \bottomrule
  \end{tabular}
\end{table}

We present only the metrics that are computed by the competition organizers in the testing stage for the clear comparison, apart from the manual metrics and PreSumm approach Table~\ref{tab:validation}.
Full performance results for all 24 metrics on validation can be found the project source page [15]. 

\subsection{Testing}

\begin{table*}[bp]
  \caption{Human Evaluation}
  \label{tab:human}
  \centering
  \begin{tabular}{l r r r r  }
    \toprule
     Method &  Adequacy &  Grammatical Correctness &  Fluency & Total Score \\
     \midrule
     $ABC$ & $3.93$ & $4.46$ & $4.32$ & $12.71$ \\
     $Hitachi$ & $4.30$ & $4.11$ & $4.00$ & $12.41$ \\
     $UEDIN$ & $1.91$ & $3.92$ & $3.31$ & $9.14$ \\
     $JU_PAD$ & $2.71$ & $3.00$ & $3.07$ & $8.78$ \\
     $Symantlytical$ & $2.35$ & $3.22$ & $3.00$ & $8.57$ \\
     $The Turing TESTament$ & $3.00$ & $2.80$ & $2.36$ & $8.16$ \\
     $Auto Minuters$ & $2.21$ & $2.75$ & $2.60$ & $7.56$ \\
     $Matus_Francesco$ & $2.29$ & $2.92$ & $2.14$ & $7.35$ \\
     $VecSumm$ & $ \bf 1.857$ & $ \bf 2.304$ & $ \bf 1.911$ & $6.19$ \\
     $Pegasus$ & $ \bf 1.250$ & $ \bf 2.607$ & $ \bf 1.786$ & $5.32$ \\
     $T5$ & $ \bf 1.107$ & $ \bf 2.571$ & $ \bf 1.732$ & $5.09$ \\
     $PreSumm$ & $ \bf 1.482$ & $ \bf 1.964$ & $ \bf 1.393$ & $4.43$ \\
    \bottomrule
  \end{tabular}
  
\end{table*}

Metrics for the test (test I and test II) sets were calculated by the organizers of the competition and are shown in Table~\ref{tab:eval} (automatic evaluation) and Table~\ref{tab:human} (human evaluation). For the testing phase we managed to include the PreSumm approach.

Among our models, VecSumm approach has the better ROUGE metrics result at the testing phase (Table~\ref{tab:eval}), which correlates with the metrics at the validation phase (Table~\ref{tab:validation}). On the contrary, Pegasus has the similar metrics as VecSumm on the validation dataset, but during the test, the ROUGE metrics dropped dramatically which may indicate a strong overfitting on train data.

\begin{table}[th]
  \caption{Automatic Evaluation}
  \label{tab:eval}
  \centering
  \begin{tabular}{l r  r   r  }
    \toprule
     Method &  Rouge 1 &  Rouge 2 &  Rouge L \\
     \midrule
     $ABC$ & $0.282$ & $0.065$ & $0.160$ \\
     $JU_PAD$ & $0.221$ & $0.047$ & $0.126$ \\
     $Hitachi$ & $0.217$ & $0.060$ & $0.116$ \\
     $Symantlytical$ & $0.217$ & $0.042$ & $0.111$ \\
     $Auto Minuters$ & $0.211$ & $0.044$ & $0.116$ \\
     $VecSumm$ & $ \bf 0.180$ & $ \bf 0.035$ & $ \bf 0.098$ \\
     $UEDIN$ & $0.173$ & $0.039$ & $0.118$ \\
     $Matus_Francesco$ & $0.170$ & $0.034$ & $0.091$ \\
     $The Turing TESTament$ & $0.157$ & $0.046$ & $0092$ \\
     $PreSumm$ & $ \bf 0.130$ & $ \bf 0.017$ & $ \bf 0.069$ \\
     $Pegasus$ & $ \bf 0.064$ & $ \bf 0.008$ & $ \bf 0.047$ \\
     $T5$ & $ \bf 0.045$ & $ \bf 0.005$ & $ \bf 0.035$ \\
    \bottomrule
  \end{tabular}
\end{table}

Among all participants in the competition, our best approach VecSumm takes 6th place according to automatic evaluation and 9th place according to human evaluation.

\section{Conclusions}

A comparative study was carried out for the task of summarization of manually transcribed audio recordings. We have explored multiple deep learning models pre-trained on large corpora to tackle the text summarization task, proposed a method for improved fine-tuning such systems and proposed VecSumm, an alternative technique that requires no pre-training. We have shown that it is possible to obtain results comparable with deep learning-based methods using VecSumm. We have also included speech recognition module in the pipeline so that we directly tackle the task of audio summarization and made all our developments open-source \cite{MTS}. 

Despite the working pipeline being set up, there are a lot of issues with the resulting transcriptions. The main issues are readability and fluency of texts, which was one the main requirements of the competition for the algorithm. Extraction of relevant facts is also weak and does not meet the quality demands of a production-ready algorithm. These issues resulted in low human scores for the final test set prediction evaluations. 

In the future, we would like to consider transcripts segmentation to fine-tune models on segmented input texts and provide summaries for extracted segments. Also, we consider to work on the postprocessing of the output of our summarization algorithms to improve coherence and fluency of the resulting minutes. For example, it is necessary to resolve coreferences in names in regard to the speaker. Also, more work should be done on making the resulting sentences short and concise, having unnecessary words and phrases removed.
% \subsection{References}

% The reference format is the standard IEEE one. References should be numbered in order of appearance, for example \cite{Davis80-COP}, \cite{Rabiner89-ATO}, \cite[pp.\ 417--422]{Hastie09-TEO}, and \cite{YourName21-XXX}.

% \subsection{Submitted files}

% Authors are requested to submit PDF files of their manuscripts. You can use commercially available tools or for instance http://www.pdfforge.org/products/pdfcreator. The PDF file should comply with the following requirements: (a) there must be no PASSWORD protection on the PDF file at all; (b) all fonts must be embedded; and (c) the file must be text searchable (do CTRL-F and try to find a common word such as ``the''). The proceedings editors (Causal Productions) will contact authors of non-complying files to obtain a replacement. In order not to endanger the preparation of the proceedings, papers for which a replacement is not provided in a timely manner will be withdrawn.

% \section{Acknowledgements}

% The ISCA Board would like to thank the organizing committees of the past INTERSPEECH conferences for their help and for kindly providing the template files. \\
% Note to authors: Authors should not use logos in the acknowledgement section; rather authors should acknowledge corporations by naming them only.

\bibliographystyle{IEEEtran}

\bibliography{mybib}

% \begin{thebibliography}{9}
% \bibitem[1]{Davis80-COP}
%   S.\ B.\ Davis and P.\ Mermelstein,
%   ``Comparison of parametric representation for monosyllabic word recognition in continuously spoken sentences,''
%   \textit{IEEE Transactions on Acoustics, Speech and Signal Processing}, vol.~28, no.~4, pp.~357--366, 1980.
% \bibitem[2]{Rabiner89-ATO}
%   L.\ R.\ Rabiner,
%   ``A tutorial on hidden Markov models and selected applications in speech recognition,''
%   \textit{Proceedings of the IEEE}, vol.~77, no.~2, pp.~257-286, 1989.
% \bibitem[3]{Hastie09-TEO}
%   T.\ Hastie, R.\ Tibshirani, and J.\ Friedman,
%   \textit{The Elements of Statistical Learning -- Data Mining, Inference, and Prediction}.
%   New York: Springer, 2009.
% \bibitem[4]{YourName17-XXX}
%   F.\ Lastname1, F.\ Lastname2, and F.\ Lastname3,
%   ``Title of your INTERSPEECH 2021 publication,''
%   in \textit{Interspeech 2021 -- 20\textsuperscript{th} Annual Conference of the International Speech Communication Association, September 15-19, Graz, Austria, Proceedings, Proceedings}, 2020, pp.~100--104.
% \end{thebibliography}

\end{document}